\documentclass[conference]{IEEEtran}
\IEEEoverridecommandlockouts
\usepackage{cite}
\usepackage{amsmath,amssymb,amsfonts}
\usepackage{algorithm}
\usepackage[noEnd=false]{algpseudocodex}
\usepackage{graphicx}
\usepackage{textcomp}
\usepackage{xcolor}

\usepackage[nolist]{acronym}
\begin{acronym}[DRAM]
    \acro{ANN}{Artificial Neural Network}
    \acro{DNN}{Deep neural network}
    \acrodefplural{DNN}{Deep neural networks}
    \acro{NN}{neural network}
    \acro{ML}{machine learning}
    \acro{DL}{deep learning}
    \acro{MLP}{Multilayer Perceptron}
    \acro{TA}{Tsetlin automaton}
    \acrodefplural{TA}{Tsetlin automata}
    \acro{TM}{Tsetlin Machine}
    \acrodefplural{TM}{Tsetlin Machines}
    \acro{CTM}{Convolutional Tsetlin Machine}
    \acrodefplural{CTM}{Convolutional Tsetlin Machines}
    \acro{CoTM}{Coalesced Tsetlin Machine}
    \acro{NLP}{Natural Language Processing}
    \acro{CelebA}{Celebrity Faces and Attributes}
    \acro{AUROC}{Area Under the Receiver Operating Characteristic Curve}
    \acro{AUPRC}{Area Under the Precision-Recall Curve}
    \acro{CAM}{Class Activation Map}
\end{acronym}

\def\BibTeX{{\rm B\kern-.05em{\sc i\kern-.025em b}\kern-.08em
    T\kern-.1667em\lower.7ex\hbox{E}\kern-.125emX}}

\begin{document}

% Transparent Multi-Label Image Classification with Convolutional Tsetlin Machines
\title{A Methodology for Transparent Logic-Based Classification Using a Multi-Task Convolutional Tsetlin Machine\\
	% {\footnotesize \textsuperscript{*}Note: Sub-titles are not captured in Xplore and should not be used} \thanks{Identify
	% applicable funding agency here. If none, delete this. }
}

\author{
	\IEEEauthorblockN{Mayur Kishor Shende}
	\IEEEauthorblockA{\textit{Dept. of ICT} \\ \textit{University of Agder}\\ Grimstad, Norway \\ mayurks@uia.no}
	\and
    
	\IEEEauthorblockN{Ole-Christoffer Granmo}
	\IEEEauthorblockA{\textit{Dept. of ICT} \\ \textit{University of Agder}\\ Grimstad, Norway \\ ole.granmo@uia.no}
	\and
    
	\IEEEauthorblockN{Runar Helin}
	\IEEEauthorblockA{\textit{Dept. of ICT} \\ \textit{University of Agder}\\ Grimstad, Norway \\ runar.helin@uia.no}
	\and
    \and[\hfill\mbox{}\par\mbox{}\hfill]
    
	\IEEEauthorblockN{Vladimir I. Zadorozhny}
	\IEEEauthorblockA{\textit{School of Computing and Information} \\ \textit{University of Pittsburgh}\\ Pittsburgh, USA}
	\and
    
	\IEEEauthorblockN{Rishad Shafik}
	\IEEEauthorblockA{\textit{School of Engineering} \\ \textit{Newcastle University}\\ Newcastle, UK \\ rishad.shafik@newcastle.ac.uk}
	% \and
	% \IEEEauthorblockA{\textit{dept. name of organization (of Aff.)} \\ \textit{name of organization (of Aff.)}\\ City, Country \\ email address or ORCID}

}

\maketitle

\begin{abstract}

The Tsetlin Machine (TM) is a novel machine learning paradigm that employs finite-state
automata for learning and utilizes propositional logic to represent patterns.
Due to its simplistic approach, TMs are inherently more interpretable than
learning algorithms based on Neural Networks. The Convolutional TM has shown comparable performance on various
datasets such as MNIST, K-MNIST, F-MNIST and CIFAR-2. In this paper, we explore
the applicability of the TM architecture for large-scale multi-channel
(RGB) image classification. We propose a methodology to generate both local interpretations 
and global class representations. The local interpretations can be used to explain the model 
predictions while the global class representations aggregate important patterns for each class. 
These interpretations summarize the knowledge captured by the convolutional clauses, 
which can be visualized as images. We evaluate our methods on MNIST and CelebA datasets, 
using models that achieve 98.5\% accuracy on MNIST and 86.56\% F1-score on CelebA 
(compared to 88.07\% for ResNet50) respectively. We show that the TM performs competitively 
to this deep learning model while maintaining its interpretability, even in large-scale 
complex training environments. This contributes to a better understanding of TM clauses 
and provides insights into how these models can be applied to more complex and diverse datasets.

\end{abstract}


\section{Introduction} \label{sec:intro}
The rapid advancement of \ac{ML} has led to the widespread adoption of deep
neural networks for tasks such as image and text classification. While these
models have achieved remarkable accuracy across a variety of benchmarks, their
reliance on complex, multi-layered architectures often results in opaque
decision-making processes and significant computational
cost~\cite{saleem2022explaining, sze2017efficient}. Consequently, neural
network-based models are often referred to as black boxes. Transparent
decision-making and interpretable models are crucial in many domains, such as
healthcare, finance, and legal systems. One reason is that \ac{ML} models are known to
replicate the biases present in the training data\cite{mehrabi2021survey}, hence, 
requiring human oversight. The biases can also be amplified, because \ac{ML} based 
decision support systems have been shown to adversely affect the decisions of
their users~\cite{glickman2025human}. A lack of transparency makes it difficult to
identify and mitigate these challenges.

In contrast, the \ac{TM}~\cite{granmo2018tsetlin} is a novel \ac{ML} paradigm
that employs finite-state machines for learning and utilizes propositional
logic to represent patterns. Unlike deep learning classifiers, which rely on
complex networks with multiple layers of nonlinear transformations, making them
difficult to interpret, \acp{TM} operate on binarized input data and generate
propositional AND-rules. This approach enhances transparency and
interpretability, as the decision-making process can be directly traced through
these logical clauses. \acp{TM} leverage simple bitwise operations, leading to
competitive accuracy across various benchmarks while significantly reducing
computational complexity and energy consumption. These
characteristics make them hardware-friendly~\cite{tunheim2024tsetlin, tm-edge,
	wheeldon2020learning}.

The \ac{TM} and its variants have been shown to achieve competitive results in
various \ac{ML} applications, such as \ac{NLP}~\cite{saha2022relational,
	rbe-rohan, human-sentiment, tm-ae, berge2019using}, classification and
regression tasks~\cite{abeyrathna2019, drop-clause, parallel-tm}, signal
processing~\cite{jeeru2025interpretable}, federated
learning~\cite{qi2025fedtmos}, and the contextual bandit problem~\cite{bandit}.
The convergence properties of \ac{TM} have been analyzed in studies such
as~\cite{jiao2021convergence, zhang2020convergence}. Granmo \textit{et
	al.}~\cite{granmo2019convolutional} introduced the \ac{CTM}, which has been
shown to achieve state-of-the-art performance for image classification on
datasets such as MNIST, K-MNIST, F-MNIST, and CIFAR-2.

However, the literature on the application of the \ac{CTM} to large-scale RGB
image datasets and the analysis of convolutional clauses is limited. With such
datasets, the number of features and the complexity of the patterns increase,
which also results in a high number of clauses. The clauses learned by the
\ac{CTM} are used as convolutional filters, which adds to the complexity of
interpretation. Furthermore, the clauses do not directly map to the input image
and also encode spatial information using thermometer encoding. These factors
make interpreting convolutional clauses non-trivial.

This paper aims to address this challenge by proposing a methodology to
generate an interpretation for the convolutional clauses, which can be used to
explain the predictions of the model. The methodology provides a summary of the 
knowledge captured by the convolutional clauses which can be visualized in the 
form of images. Our approach is designed to efficiently
generate a local interpretation for the clauses. We also present a strategy to
generate a class-wise global representation, which aggregates the important
features for a class. We demonstrate that \acp{TM} can maintain their hallmark
interpretability and transparency while achieving classification accuracy
comparable to deep learning models, even in large-scale, complex training
environments.

The remainder of this paper is organized as follows.
Section~\ref{sec:background} provides background on Tsetlin Automata, Tsetlin
Machines, and their convolutional extensions. Section~\ref{sec:methodology}
details our proposed methodology for interpreting convolutional clauses,
including both local and global interpretation strategies.
Section~\ref{sec:exp} describes the experimental setup, datasets, and
evaluation metrics. Section~\ref{sec:result} presents and analyzes the results.
Finally, Section~\ref{sec:conclusion} concludes the paper and discusses
directions for future work.

% What is the problem?

% Why is it hard? DL explainability is approx. natively intepretable methods struggle with non symbolic high dimentional
% data such as images and Natural language. TM is promising new approach, concerning images and text, unimodel single
% task problems. there is lack of methodology for multitask and multimodel, global and local intepretability. THere are
% no methods for ConvTM. make the clauses understandable for end users. ECG example

% What are the key components of my approach nd results? A way to combine the patches explain local and global Key
% finding for mnist and celebrity: ability to deal with synonyms and non related words the intensity of vidualisation
% shows cnfidence int the classification.

\section{Background} \label{sec:background}

\subsection{Tsetlin Automaton}

\begin{figure}[htbp]
	\begin{center}
		\includegraphics[width=0.95\linewidth]{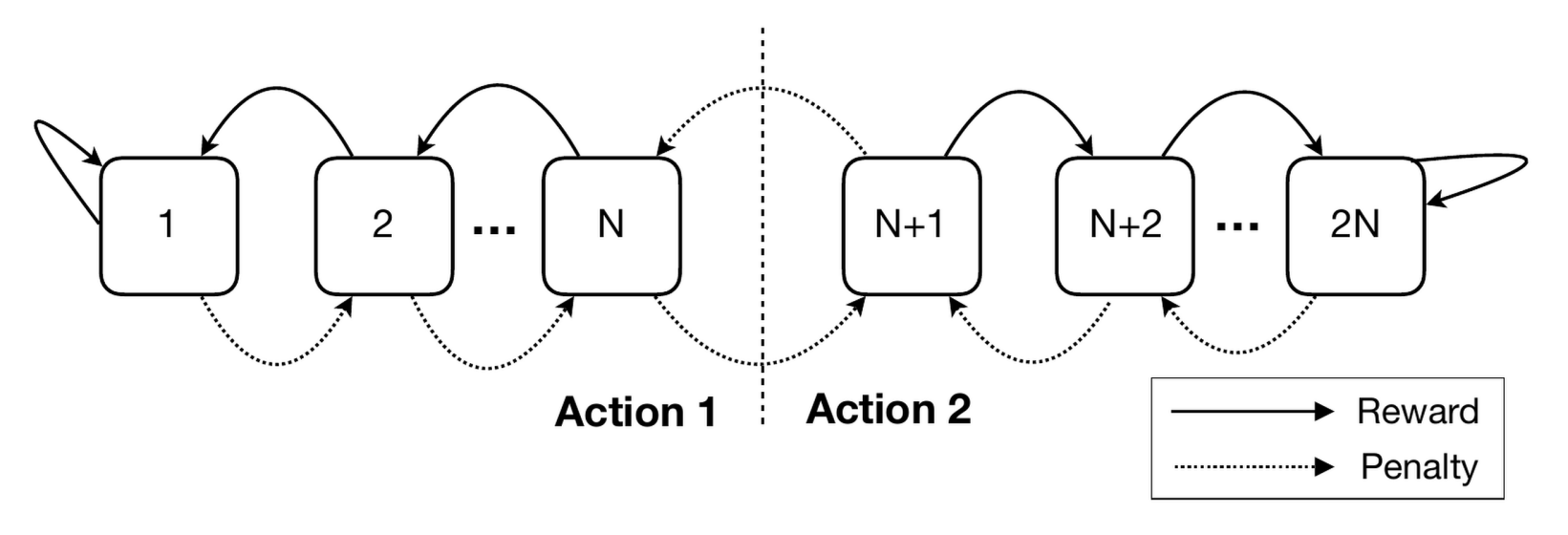}
	\end{center}
	\caption{A \ac{TA} with $2N$ states and two actions}
	\label{fig:ta}
\end{figure}

The \ac{TA} is a type of learning automaton designed to learn optimal actions
in stochastic environments. Based on environment feedback, it either receives a
reward (state increment) or a penalty (state decrement). The reward has an
associated probability that can change over time. Fig.~\ref{fig:ta} shows a
\ac{TA} consisting of $2N$ states, where states $1$ to $N$ correspond to the
Exclude action (Action 1), and states $N+1$ to $2N$ correspond to the Include action (Action 2).

\subsection{Tsetlin Machine}

The \ac{TM} is a collection of \acp{TA} organized into multiple teams,
responsible for learning different patterns in the data. Each feature in the
input data is associated with a \ac{TA}, and a team of these \acp{TA}
collectively forms a \textit{clause}. Each \textit{clause} votes for a class
label, and these votes are aggregated to calculate the predicted class label.
Formally, let $X = \{x_0, x_1, x_2, \ldots\}$ be the set of binarized input
features, and $C$ be the set of \textit{clauses}. The original features $X$ are
combined with the negated features to form the set of literals
%$L = X \cup \neg
%	X$, which are used as input to the \ac{TM}. A \textit{clause} $c_i \in C$ is
$L = X \cup
	\{\lnot x_0, \lnot x_1, \lnot x_2, \ldots\}$, which are used as input to the \ac{TM}. A \textit{clause} $c_i \in C$ is
defined as the conjunction of a subset of these literals as shown in Eq.~\ref{eq:clause}:
\begin{equation}
	c_i = \bigwedge_{f \in L_I} f,
	\label{eq:clause}
\end{equation}
where $L_I \subseteq L$ is the set of literals included. 

Half of the clauses are assigned
positive polarity ($C^+$), i.e., they vote for the class, and the other half
are assigned negative polarity ($C^-$), i.e., they vote against the class. A
\textit{clause} is active if the included literals in the clause match the
input features. In the weighted version of the \ac{TM}, each clause $k$ also has
weights, $w_k$, for each class. The polarity of $w_k$ determines if the clause is of type positive or negative polarity. The sum of the weights
of the \emph{True} (active) clauses for a class is defined as the \textit{class sum}, $v$.

\begin{equation}
	v(X) = \sum_{k=1}^{n_+} w_k C^+_k(X) + \sum_{k=1}^{n_-} w_k C^-_k(X)
	\label{eq:class_sum}
\end{equation}
The \textit{class sums} for each class dictate the prediction of the \ac{TM}.

For multi-class classification problems, the class with the highest
\textit{class sum} is selected as the predicted class:
\begin{equation}
	\hat{y} = \arg\max_{m \in M} v_m(X)
	\label{eq:multi_class_prediction}
\end{equation}

For multi-label
classification problems, where each input can have multiple class labels, the
classes with positive \textit{class sums} are selected as the predicted
classes:
\begin{equation}
	\hat{y} = \{m \in M \mid v_m(x_i) > 0\}
	\label{eq:multi_label_prediction}
\end{equation}

During learning, the \ac{TM} uses Type I (a and b) and Type II feedback
mechanisms to calculate the probabilities for updating the states of each
\ac{TA}. These probabilities depend on the \textit{class sum}, as well as the
hyperparameters specificity ($s$) and target ($T$). Type I(a) feedback
reinforces true positive samples, while Type I(b) feedback penalizes false
negative predictions. Type II feedback is used to correct false positive
predictions. A more detailed explanation of the \ac{TM} and its learning
algorithm can be found in \cite{granmo2018tsetlin}.

\subsection{Convolutional Tsetlin Machine}

\begin{figure}[htbp]
	\begin{center}
		\includegraphics[width=0.99\linewidth]{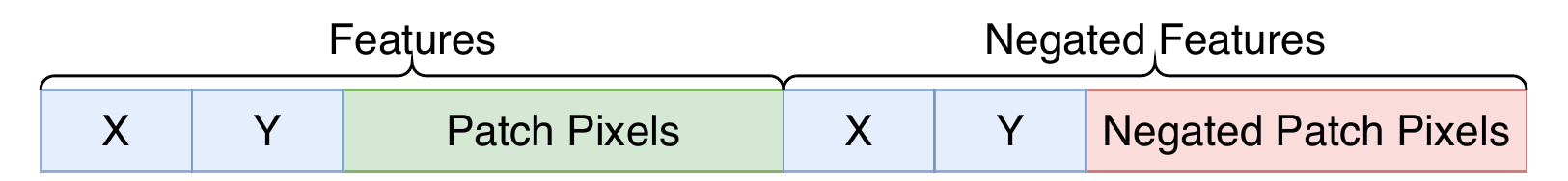}
	\end{center}
	\caption{Structure of a convolutional clause}
	\label{fig:conv_clause_structure}
\end{figure}

The \ac{CTM} is an extension of the \ac{TM} that learns smaller $W \times W$
convolutional filters. Each clause is treated as a location-aware convolutional
filter. Therefore, the set of literals $L$ contains the image patch ($W \times
	W \times Z$ pixels) and binary-encoded coordinates of the patch. The encoding
of the coordinates is done using the thermometer encoding scheme~\cite{BuckmanRRG18}, which allows
the clauses to have a range of possible locations rather than a single
location. The structure of the convolutional clause is shown in
Fig.~\ref{fig:conv_clause_structure}. Similar to the standard \ac{TM}, the
\ac{CTM} also requires the input to be binarized. This binarization is
typically done using an image thresholding method or thermometer encoding. The
final set of literals $L$ can then be defined as $L \in \{0, 1\}^{(W \times W
	\times Z + B_x + B_y) \times 2}$, where $B_x$ and $B_y$ are the literals
representing the binary-encoded coordinates. The multiplication by 2 is done to include the negated features.

The learning algorithm for the \ac{CTM} is similar to the standard \ac{TM}.
However, the main difference is in the way the clause activations are
calculated. In the \ac{CTM}, each clause is matched with all the patches in the
input image. This gives a set of clause activations, one potential activation for each position ($B_x
	\times B_y$). These are aggregated using the \textit{OR} operator to get the
final clause activation. One of these patches is finally selected at random when the
clause receives feedback. A detailed explanation of the \ac{CTM} can be found
in \cite{granmo2019convolutional}.

\subsection{Coalesced Convolutional Tsetlin Machine}

In the standard \ac{TM} architecture, each class has a separate set of clauses.
This means that if there are any common patterns between the classes, they need
to be learned separately for each class. The
\ac{CoTM}~\cite{glimsdal2021coalesced} addresses this by combining all the
clauses into a single set and sharing them across all the classes. Each clause
has a different set of weights for each class. In this approach, a clause can
have multiple polarities, i.e., it can be positive for one class and negative
for another class. This allows the \ac{CoTM} to learn complex patterns with
fewer clauses.

\section{Methodology} \label{sec:methodology}

\subsection{Local Interpretation}

The clauses learned by a \ac{TM} represent sub-patterns in the dataset.
Individual clauses cannot be used to reason about model predictions without
considering their interactions. The combination of these clauses can create
meaningful patterns, which can be used to interpret the model. Thus, by
analyzing the activated clauses for a given input, we can generate a
visual representation that highlights the patterns responsible for the prediction.
This representation, generated using the activated clauses, is the local
interpretation of the model.

In the case of the \ac{CTM}, clauses also have a spatial component, which can
be used to determine the location of the activated clause.
Algorithm~\ref{alg:local_interpretation} shows the procedure for generating the
local interpretation for a given input. The algorithm takes the trained \ac{TM}
model, the binarized image, and the number of channels in the unbinarized image
as input. The clauses and their weights are extracted from the model. Since the
\ac{TM} is trained on binarized input, the literals are also in binarized form
and need to be converted back. This process depends on the binarization scheme
used and can vary from application to application. In the algorithm, this is
represented by the \texttt{unbinarize()} function. Once the literals are
unbinarized, we match each positive polarity clause with each patch in the
input image. If the clause matches, i.e., the clause is active for the patch,
we place the unbinarized literals of the clause at the location of the patch.
This creates an intermediate expansion of the clause, which is similar to the
deconvolution of the clause. This intermediate expansion is then multiplied by
the weight of the clause. The local interpretation is then obtained by
subtracting the negative interpretation from the positive interpretation.

\begin{algorithm}[htbp]
	\caption{Local Interpretation for Convolutional \ac{CoTM}}
	\label{alg:local_interpretation}
	\begin{algorithmic}[1]
		\Require Trained \ac{TM} model $\mathcal{M}$, Binarized image $\mathbf{X} \in \{0, 1\}^{N \times M \times Z_b}$, Number of channels in unbinarized image $Z$
		\Ensure Output $I \in \mathbb{Z}^{N \times M \times Z}$
		\State $\mathbf{C} \leftarrow \mathcal{M}.\text{number\_of\_clauses}$
		\State $\mathbf{P} \leftarrow \mathcal{M}.\text{number\_of\_patches}$
		\State $\mathbf{W} \leftarrow \mathcal{M}.\text{get\_clause\_weights}()$
		\State $\mathbf{L}^+, \mathbf{L}^- \leftarrow \mathcal{M}.\text{get\_literals}()$
		\State $\mathbf{L}^+, \mathbf{L}^- \leftarrow \text{unbinarize}(\mathbf{L}^+, \mathbf{L}^-)$
		\State Initialize $\mathbf{I}^+, \mathbf{I}^- \leftarrow \text{Zeros}((N, M, Z))$
		\For{each $c \leftarrow 1, \ldots, C$}
		\State Initialize $\text{tempI}^+, \text{tempI}^- \leftarrow \text{Zeros}((N, M, Z))$
		\If{$W_c > 0$} \Comment{Positive polarity clause}
		\For{each $p \leftarrow 1, \ldots, P$}
		\State Let $x_p \leftarrow \text{get\_patch}(X, p)$
		\State $m, n \leftarrow \text{get\_coordinates}(X, p)$
		\If{$C_c \land x_p = C_c$} \Comment{$C_c$ matches patch $x_p$}
		\State $\text{tempI}^+_{m,n} \leftarrow \text{tempI}^+_{m,n} + L^+_{c}$
		\State $\text{tempI}^-_{m,n} \leftarrow \text{tempI}^-_{m,n} + L^-_{c}$
		\EndIf
		\EndFor
		\EndIf
		\State $I^+ \leftarrow W_c \times \text{tempI}^+$
		\State $I^- \leftarrow W_c \times \text{tempI}^-$
		\EndFor
		\State $I \leftarrow I^+ - I^-$
		\State \Return $I$
	\end{algorithmic}
\end{algorithm}

\subsection{Global Class Representation}

\begin{figure*}[htbp]
	\begin{center}
		\includegraphics[width=0.95\linewidth]{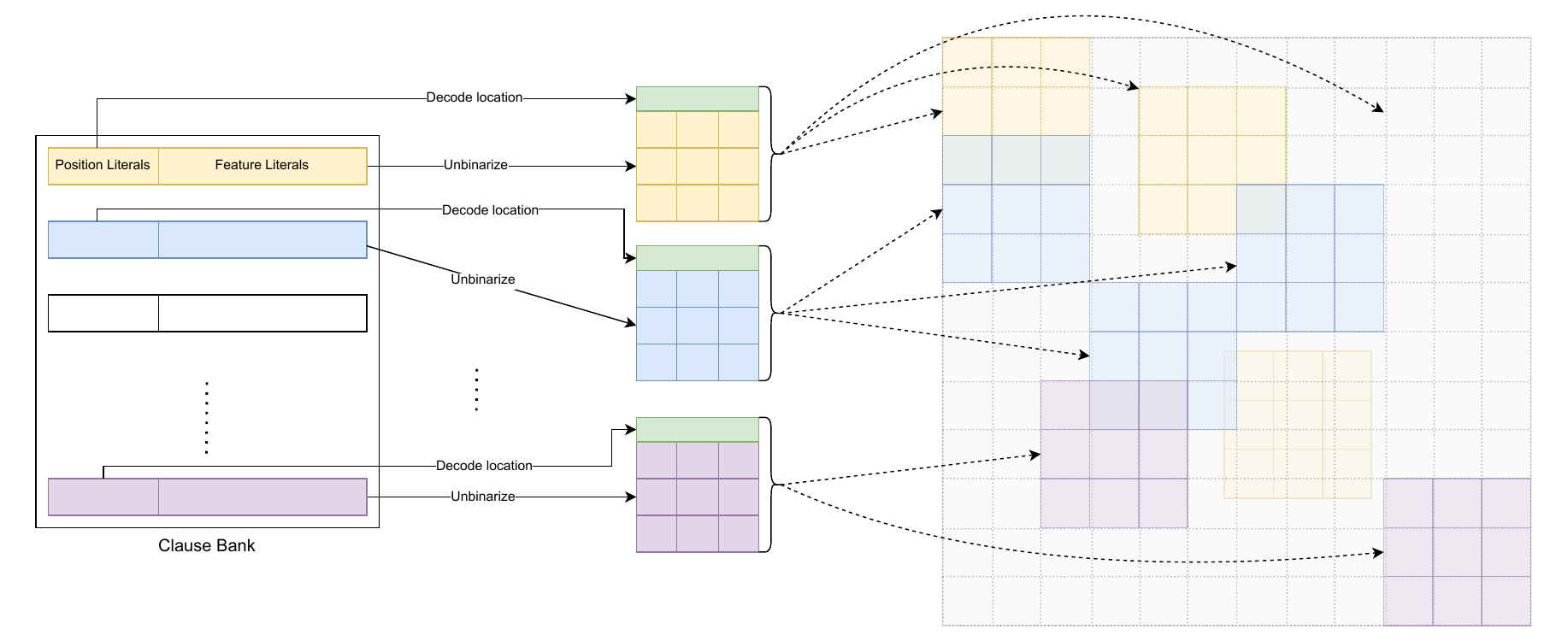}
	\end{center}
	\caption{Combining convolutional clauses for interpretation}
	\label{fig:methodology}
\end{figure*}

The local interpretation represents the patterns activated for a given input.
However, it does not show all the patterns that are important for a class. To
obtain a global representation of the class, we need to analyze all the
positive polarity clauses for a class. Since the positive polarity clauses
learn the patterns favoring a class, a combination of these clauses can be used
to represent this class. This is trivial in the case of the standard \ac{TM},
where the clauses directly represent the input features. However, in the case
of \ac{CTM}, the clauses do not map directly to
the input image. Also, the clauses contain spatial information, encoded using
thermometer encoding. This allows the clause to match a range of locations,
rather than being restricted to a single location. However, this also makes
interpreting the clause challenging because it could be activated at any
location in the range. Due to this, it is not possible to obtain an exact
representation of the class. Instead, we can approximate the class
representation using the method of patch counting. The process is summarized in
Fig.~\ref{fig:methodology}.

\textbf{Patch Counting}: By analyzing the location literals of a clause, it can
be seen that a clause tends to specialize towards a specific region in the
image. Patch count refers to the frequency of a clause activation at each location. As the model is trained, the frequency for a certain range of locations becomes higher compared. This means that the clause focuses on this region of the input image, and the normalized frequency can be used as a relative weight for each location.

\textbf{Thermometer Decoding}: The thermometer encoding scheme is used to
encode the coordinates of a patch. Let $B$ be the total number of values
(coordinates, in this case) to encode, and $x$ be the current value, then the
thermometer encoding is a binary vector of length $B-1$, and is given by
Eq.~\ref{eq:thermometer_encoding}.

\begin{equation}
	\mathcal{T}(x, B) = [\underbrace{1, 1, \ldots, 1}_{x \text{ ones}}, \underbrace{0, 0, \ldots, 0}_{(B-1-x) \text{ zeros}}]
	\label{eq:thermometer_encoding}
\end{equation}

When decoding the location literals in a clause, the only bit which matters is
the rightmost set bit. The clause will then match all the positions greater
than this set bit. An example is shown in Table~\ref{tab:therm_decode_eg}. Here
the rightmost set bit is at position 3 (assuming 0-indexing), which means the clause matches
positions 4, 5, and 6.

\begin{table}[htbp]
	\caption{Thermometer Decoding Example}
	\centering
	\begin{tabular}{|c|c|c|c|}
		\hline
		\textbf{Clause} & \multicolumn{2}{|c|}{\textbf{Input $X$}} & $\mathbf{C \land X = C}$                   \\ \cline{2-3}
		($C$)           & Encoded                                  & Value                    & (Clause Match?) \\ \hline
		                & 000000                                   & 0                        & 0               \\ \cline{2-4}
		                & 100000                                   & 1                        & 0               \\ \cline{2-4}
		                & 110000                                   & 2                        & 0               \\ \cline{2-4}
		100100          & 111000                                   & 3                        & 0               \\ \cline{2-4}
		                & 111100                                   & 4                        & 1               \\ \cline{2-4}
		                & 111110                                   & 5                        & 1               \\ \cline{2-4}
		                & 111111                                   & 6                        & 1               \\ \hline
	\end{tabular}
	\label{tab:therm_decode_eg}
\end{table}

Algorithm~\ref{alg:global_repr} describes the algorithm for generating the
global class representation for a class $l$. The patch counting strategy
described above is used during model training and is stored in the model. The
\texttt{get\_patch\_weights()} function is used to retrieve the frequencies and
normalize them. The \texttt{decodePositionLiterals()} function implements the
thermometer decoding scheme and returns the coordinates for the positive and
negative literals separately. Similar to the local interpretation, the positive
and negative literals are unbinarized using the \texttt{unbinarize()} function.
The algorithm then iterates over all positive polarity clauses. For each
position calculated by the \texttt{decodePositionLiterals()} function, the
literals are placed at the corresponding position, weighted by the patch weight
$V$. This creates an intermediate representation for each clause. The final
global class representation is obtained by aggregating the intermediate
representations for all the clauses, weighted by the clause weights $W$.

\begin{algorithm}[htbp]
	\caption{Global Class Representations for Convolutional \ac{CoTM}}
	\label{alg:global_repr}
	\begin{algorithmic}[1]
		\Require Trained \ac{TM} model $\mathcal{M}$, Number of channels in unbinarized image $Z$, Class index $l \in \{1, \ldots, \text{labels}\}$

		\Ensure Output $I \in \mathbb{Z}^{N \times M \times Z}$
		% \Ensure Spatial interpretation of clauses $\mathbf{I} \in \mathbb{R}^{N \times H \times W}$

		\State $\mathbf{C} \leftarrow \mathcal{M}.\text{number\_of\_clauses}$
		\State $\mathbf{P} \leftarrow \mathcal{M}.\text{number\_of\_patches}$
		\State $\mathbf{W} \leftarrow \mathcal{M}.\text{get\_clause\_weights}()$
		\State $\mathbf{V} \leftarrow \mathcal{M}.\text{get\_patch\_weights}()$
		\State $\mathbf{L}^+, \mathbf{L}^- \leftarrow \mathcal{M}.\text{get\_literals}()$
		\State $\mathbf{K}^{+}, \mathbf{K}^{-} \leftarrow \text{decodePositionLiterals}(\mathbf{L}^+, \mathbf{L}^-)$
		\State $\mathbf{L}^+, \mathbf{L}^- \leftarrow \text{unbinarize}(\mathbf{L}^+, \mathbf{L}^-)$

		\State Initialize $\mathbf{I}^+, \mathbf{I}^- \leftarrow \text{Zeros}((N, M, Z))$

		\For{each $c \leftarrow 1, \ldots, C$}
		\State Initialize $\text{tempI}^+, \text{tempI}^- \leftarrow \text{Zeros}((N, M, Z))$
		\If{$W_c > 0$} \Comment{Positive polarity clause}
		\For{each $p \leftarrow 1, \ldots, P$}
		\State Let $x_p \leftarrow \text{get\_patch}(X, p)$
		\State $m, n \leftarrow \text{get\_coordinates}(X, p)$
		\If{$m, n \in K^+_c$}
		\State $\text{tempI}^+_{m,n} \leftarrow \text{tempI}^+_{m,n} + L^+_{c} \times V_{c,m, n}$
		\EndIf
		\If{$m, n \in K^-_c$}
		\State $\text{tempI}^-_{m,n} \leftarrow \text{tempI}^-_{m,n} + L^-_{c} \times V_{c,m, n}$
		\EndIf
		\EndFor
		\EndIf
		\State $I^+ \leftarrow W_c \times \text{tempI}^+$
		\State $I^- \leftarrow W_c \times \text{tempI}^-$
		\EndFor
		\State $I \leftarrow I^+ - I^-$
		\State \Return $I$
	\end{algorithmic}
\end{algorithm}

\section{Experimental Setup} \label{sec:exp}

\subsection{MNIST}

The MNIST dataset \cite{lecun1998gradient} is a widely used benchmark dataset
in the field of \ac{ML}, particularly for image classification tasks. It
consists of grayscale images of handwritten digits with 10 class labels. Each
image has $28 \times 28$ pixels. The dataset is split into 60,000 training
samples and 10,000 test samples. Since the input to the \ac{TM} needs to be
binarized, we use a single-value thresholding method to binarize the images.

A multi-class Convolutional \ac{CoTM} model was trained with 2500 clauses,
threshold($T$) of 3125, specificity($s$) of 10, and patch size $10 \times 10$. With these
hyperparameter, the model achieved an accuracy of 98.5\% on the test set. These
results are comparable to the state-of-the-art and are summarized in
Table~\ref{tab:mnist_results}.

\begin{table}[htbp]
	\caption{Hyperparameters and Classification accuracy for the MNIST dataset}\label{tab:mnist_results}
	\centering
	\begin{tabular} {l l l l l}
		\hline
		\textbf{Number of Clauses} & \textbf{T} & \textbf{s} & \textbf{Patch size} & \textbf{Accuracy} \\
		\hline
		2500                       & 3125       & 10         & 10                  & 98.5\%            \\
		\hline
	\end{tabular}
\end{table}

\subsection{Celebrity Faces and Attributes (CelebA)}

\begin{figure}[htbp]
	\begin{center}
		\includegraphics[width=0.75\linewidth]{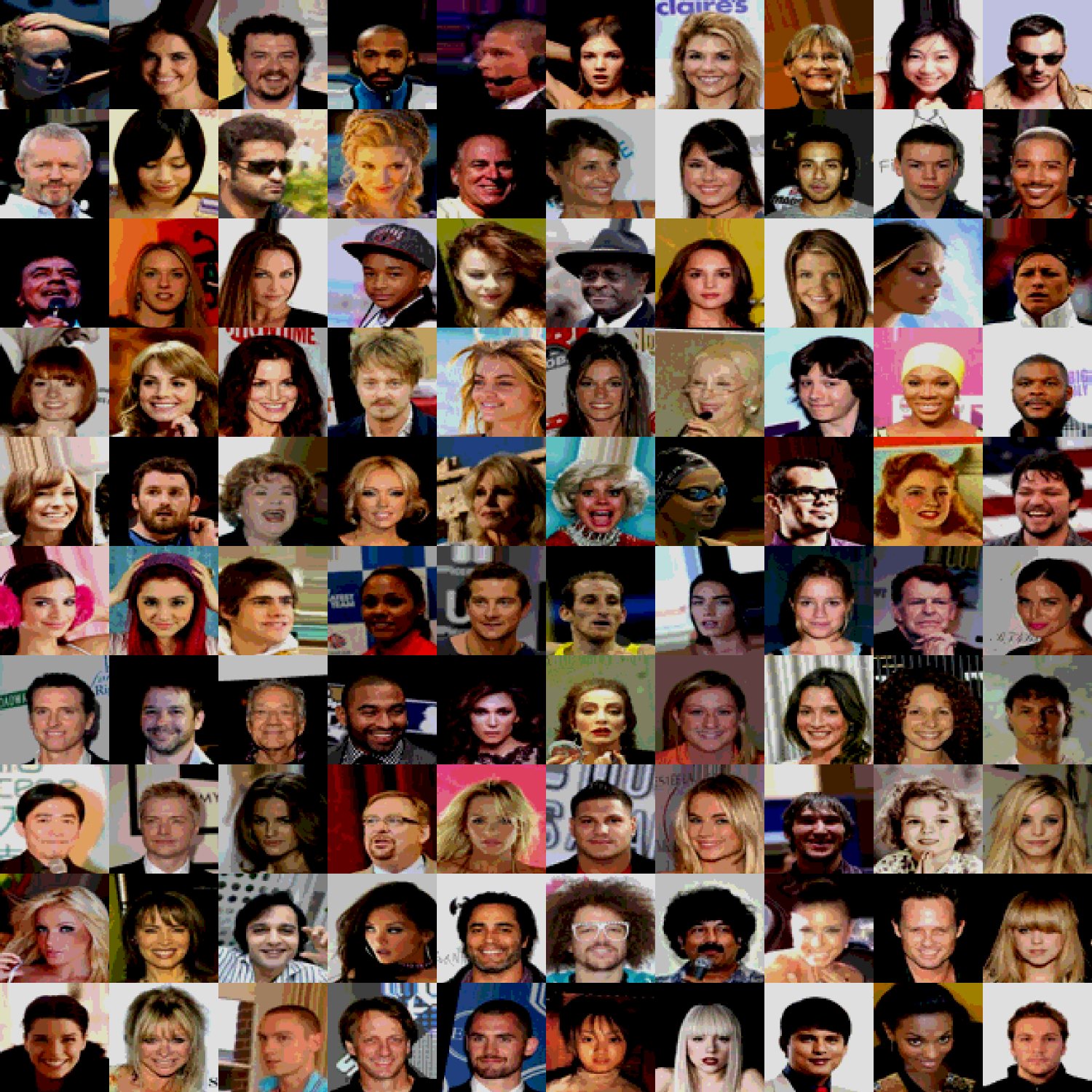}
	\end{center}
	\caption{Random samples from the \ac{CelebA} dataset.}
	\label{fig:celeba_samples}
\end{figure}

\begin{figure}[htbp]
	\begin{center}
		\includegraphics[width=0.95\linewidth]{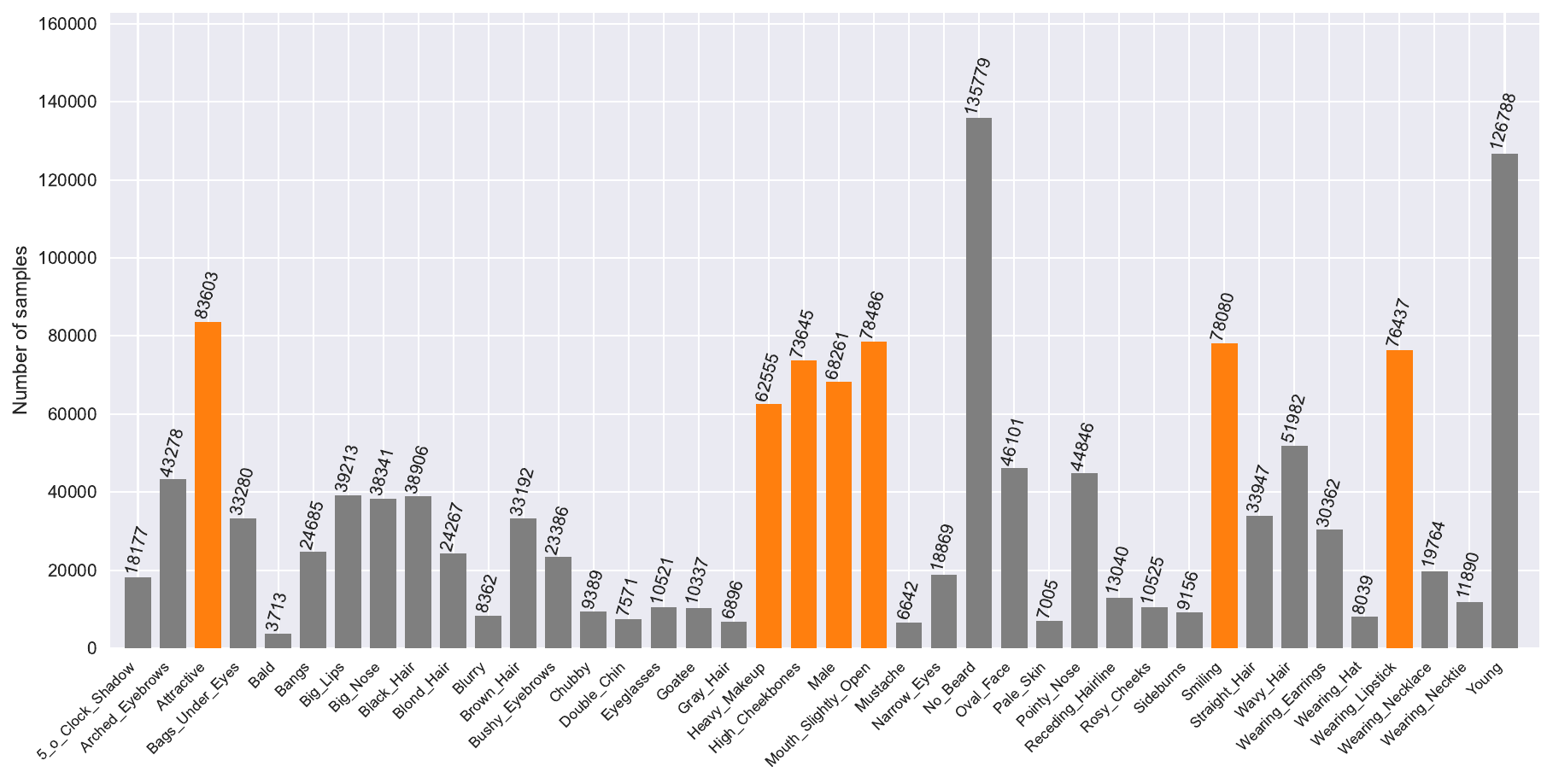}
	\end{center}
	\caption{The distribution of the classes in the \ac{CelebA} dataset. The orange bars indicate the classes selected for the experiment.}
	\label{fig:celeba_distribution}
\end{figure}

The \ac{CelebA} dataset \cite{liu2015faceattributes} is a large-scale dataset
with over 200,000 images of celebrity faces. Each image in the dataset has
3-color channels (RGB) and annotated with 40 different facial attribute labels
such as ``Smiling'', ``Male'', ``Attractive'', etc. Since each image is
associated with more than one class label, this is a multi-label/multi-output
classification problem. Fig.~\ref{fig:celeba_distribution} shows the
distribution of samples for each class in the dataset, demonstrating that the
dataset is highly imbalanced. For the purposes of this paper, a balanced subset
of 7 classes was selected, where the faces are aligned and scaled down to $64
	\times 64$ pixels. Fig.~\ref{fig:celeba_samples} shows some samples from the
dataset. The images were binarized using the thermometer encoding scheme
\cite{gronningsaeter2024optimized} with 8 levels. The binarization procedure
was applied separately for each color channel.

A hyperparameter search was performed to find the optimal hyperparameters for
the model. Since the dataset is multi-label, the Multi-output Convolutional
\ac{CoTM} was used in this case. The model was trained with 25000 clauses,
threshold($T$) of 40000, specificity($s$) of 27, patch size of $3\times 3$, and $q$ of 4.
With these hyperparameters, the model achieved an F1 score of 86.56\% on the
test set.

To compare the results against a deep learning model, a standard
ResNet50\cite{he2016deep} model was trained on the same balanced subset of the
dataset. The input images were normalized to have a mean of 0 and standard
deviation of 1. The final layer of the ResNet50 model was replaced with a fully
connected layer with 7 neurons, corresponding to the 7 classes in the dataset.
The binary cross-entropy loss with Adam optimizer was used for training. The
ResNet50 model achieved a test F1 score of 88.07\%.
Table~\ref{tab:celeba_results} reports different comparison metrics for the
\ac{CoTM} and the ResNet50 model.

\begin{table}[htbp]
	\caption{Comparing metrics for the CelebA dataset}\label{tab:celeba_results}
	\centering
	\begin{tabular} {lllll}
		\hline
		\textbf{Model}  & \textbf{Accuracy} & \textbf{F1 Score} & \textbf{AUROC} & \textbf{AUPRC} \\
		\hline
		Conv. \ac{CoTM} & 86.50\%            & 86.56\%           & 93.60\%         & 93.47\%        \\
		\hline
		ResNet50        & 88.82\%           & 88.07\%           & 94.87\%        & 94.83\%        \\
		\hline
	\end{tabular}
\end{table}

The reported metrics in the Table~\ref{tab:celeba_results} were calculated by
averaging the metric for individual classes. To calculate the \ac{AUROC} and
\ac{AUPRC}, the model needs to output the probability or confidence score for
each class. In the case of \ac{TM}, the model outputs the predicted class
labels and the class sums for each class. These class sums reflect how
confident the model is about the prediction, and thus can be converted to
probability scores \cite{helin2025uncertainty}, using Eq.~\ref{eq:class-prob}.

\begin{equation}\label{eq:class-prob}
    P(y) = \frac{1}{2}\left(1 + \frac{v(X)}{T}\right),
\end{equation}

where $P(y)$ is the probability score for a class, $v(X)$ is the class sum for the input $X$, and $T$ is the hyperparameter \textit{target value} of the \ac{TM}.

\section{Results and Discussion} \label{sec:result}

\subsection{Local Interpretation}

\subsubsection{MNIST}

\begin{figure}[htbp]
	\begin{center}
		\includegraphics[width=0.95\linewidth]{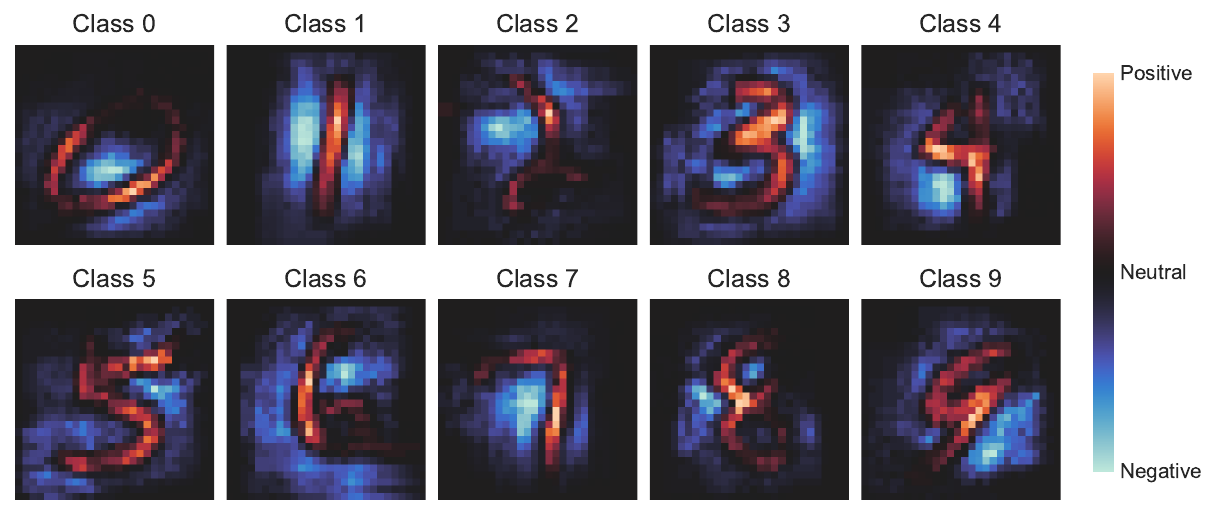}
	\end{center}
	\caption{The local interpretation for a random sample for each class from the MNIST dataset. The positive and negative literals are shown in red and blue respectively. The black region indicates the region of non importance. }
	\label{fig:mnist_local_interpretation}
\end{figure}

Fig.~\ref{fig:mnist_local_interpretation} shows the local interpretations for
the MNIST dataset. Since the dataset is grayscale, it is possible to visualize
the positive literal patterns and negative literal patterns separately. The
output of the Algorithm \ref{alg:local_interpretation}, $I \in \mathbb{Z}^{N
		\times M \times Z}$, is unbounded. In order to obtain a correct visualization,
$I$ was scaled using the Eq.\ref{eq:norm}. This normalizes the output to
$[-1, 1]$. The negative values in $I_{norm}$ corresponds to negative literals
(blue), and the positive values correspond to positive literals (red). The
intensity of the color is proportional to the literals importance, i.e., the
frequency of a literal being included in a clause.

\begin{equation}
	I_{norm} = \begin{cases}
		\frac{-v}{I_{min}} & v < 0    \\
		\frac{v}{I_{max}}  & v \geq 0
	\end{cases}
	\quad \forall v \in I
	\label{eq:norm}
\end{equation}

\subsubsection{CelebA}

\begin{figure}[htbp]
	\begin{center}
		\includegraphics[width=0.95\linewidth]{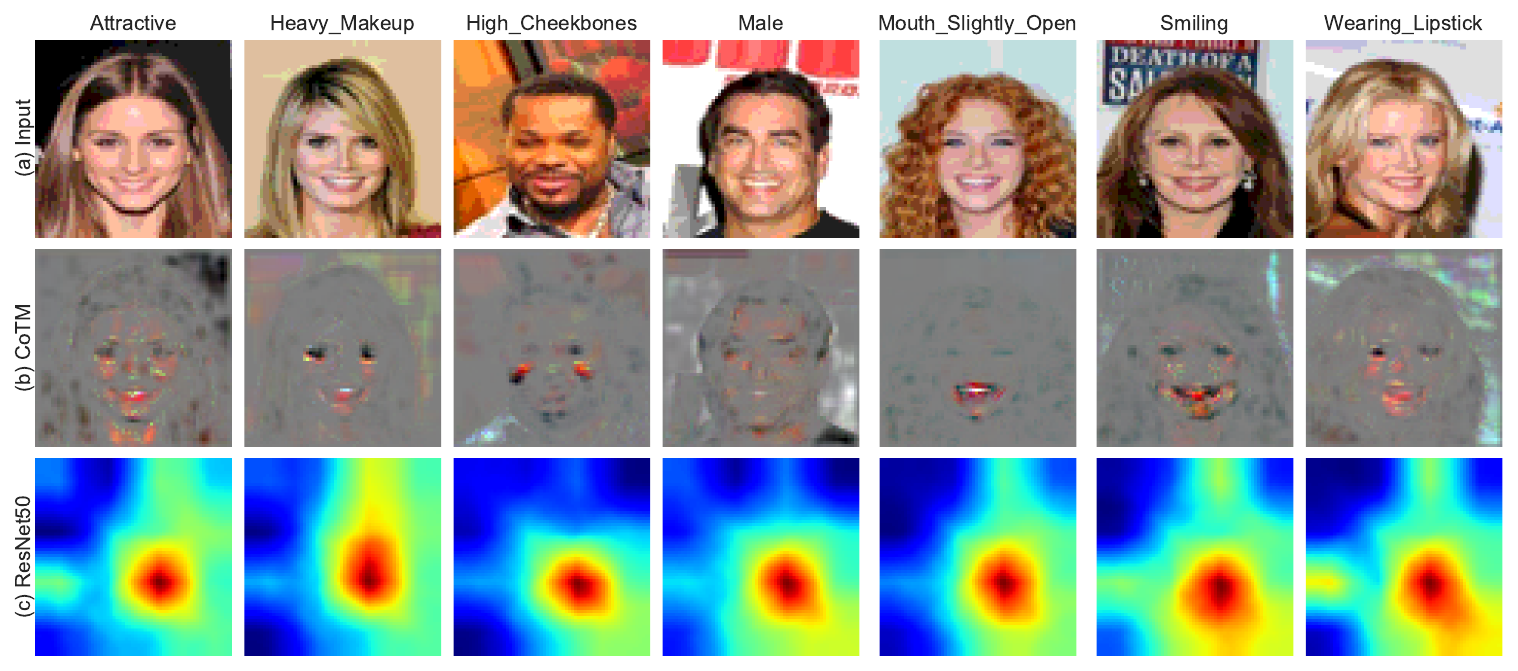}
	\end{center}
	\caption{(a) The input images from the CelebA dataset. (b) The local interpretation generated using the Convolutional \ac{CoTM} model. (c) \acp{CAM} generated using the FullGrad method using the ResNet50 model. }
	\label{fig:celeba_local_interpretation}
\end{figure}

Fig.~\ref{fig:celeba_local_interpretation} shows the local interpretation for the \ac{CelebA} dataset, and compares it with the interpretation generated using the FullGrad~\cite{srinivas2019full} method on the ResNet50 model. FullGrad is a gradient-based \ac{NN} interpretation method, that aggregates both input gradients and bias gradients across all layers of the network to create a comprehensive saliency maps. In the context of this paper, FullGrad serves as a representative baseline for neural network interpretability, generating heatmaps that indicate regions of importance for ResNet50's classification decisions.

Since the dataset contains colored images, the interpretations generated for the \ac{TM} model also have multiple color channels. Similar to the MNIST dataset, normalization was applied to the output of the Algorithm \ref{alg:local_interpretation}, separately for each color channel. Due to the interpretable nature of the \ac{TM} architecture, the local interpretations can be directly traced back to the actual pixels in the image. Because of this, the local interpretation is able to recreate important patterns in the image. In contrast, the FullGrad method generates a heatmap, indicating the region of importance.

\subsection{Global class representation}

\subsubsection{MNIST}

\begin{figure}[htbp]
	\begin{center}
		\includegraphics[width=0.95\linewidth]{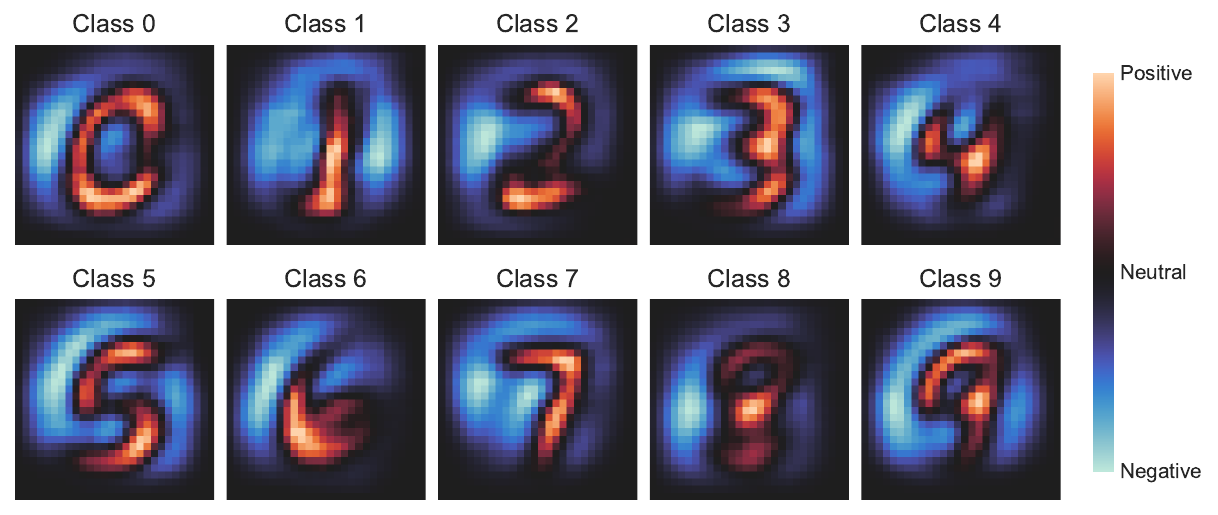}
	\end{center}
	\caption{The global class representation for the MNIST dataset.
		The red and blue regions indicates the positive and negative literals respectively.
	}
	\label{fig:mnist_global_interpretation}
\end{figure}

Fig.~\ref{fig:mnist_global_interpretation} shows the representation for each
class in the MNIST dataset. Similar to the local interpretation, the output
calculated by the algorithm is unbounded, and is normalized using
Eq.\ref{eq:norm}. The red region together forms the pattern for the
class. The blue region corresponds to the negative literals, which indicates
regions that should not be present. The black region indicates absence of
literals in any of the clauses, and thus does corresponds to ``don't
	care'' region.

\subsubsection{CelebA}

\begin{figure}[htbp]
	\begin{center}
		\includegraphics[width=0.95\linewidth]{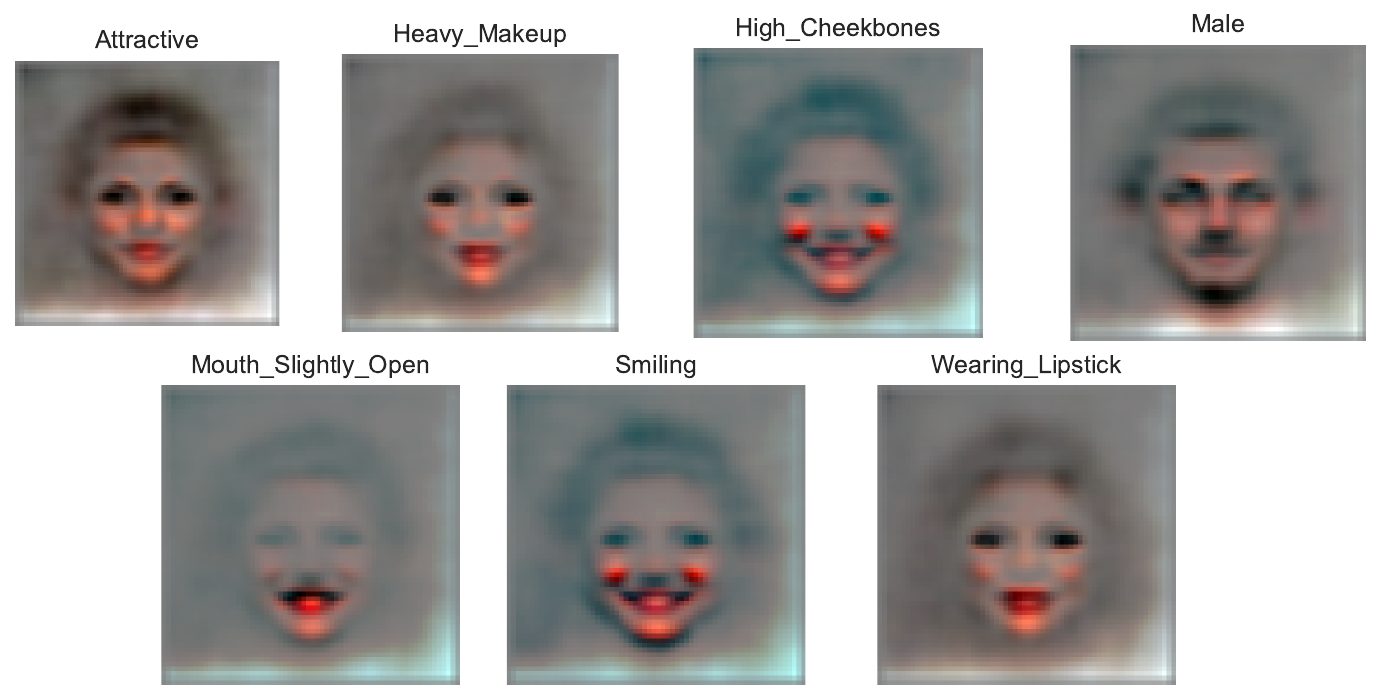}
	\end{center}
	\caption{The global class representation for the CelebA dataset.}
	\label{fig:celeba_global_interpretation}
\end{figure}

Fig.~\ref{fig:celeba_global_interpretation} shows the global class
representation for the \ac{CelebA} dataset, which is created by normalizing the
output of the Algorithm~\ref{alg:global_repr}. Since the input images in this
case were RGB, the generated representations is also RGB. This aggregates all
the patterns important for each class, and also shows the difference between
the patterns for the classes. If some classes have similar patterns and are highly
correlated, the global representation will also be similar. This can be seen
with the High Cheekbones and Smiling classes, which share most of the patterns.
The Male class is negatively correlated with all the other classes,
and thus has most distinctive patterns.

\begin{figure}[htbp]
	\begin{center}
		\includegraphics[width=0.95\linewidth]{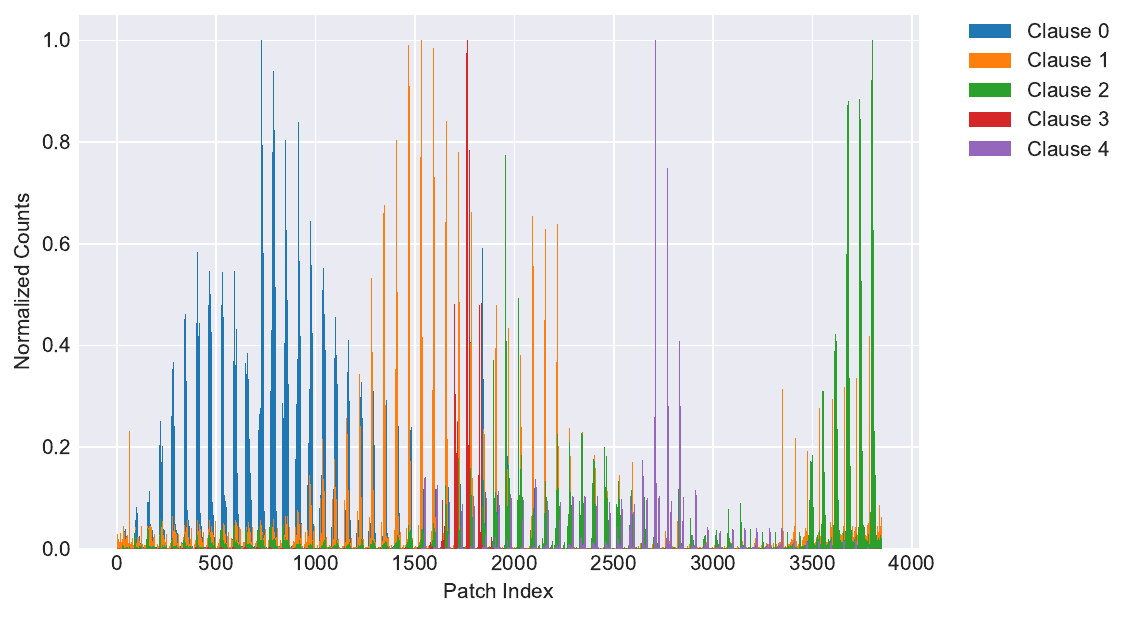}
	\end{center}
	\caption{Histogram showing the patch counts for some of the learned clauses.}
	\label{fig:pc_1}
\end{figure}

To generate the global class representation, we introduced the patch counting
mechanism, which counts the frequency of a clause activating at each possible
locations. Plotting this frequency as a histogram reveals that most of the
clauses tend to specialize towards a specific region in the image.
Fig.~\ref{fig:pc_1} show the patch counts for two random clauses learned by
the \ac{CoTM} model trained on the \ac{CelebA} dataset.

\subsection{Hyperparameter q}

\begin{figure*}[htbp]
	\begin{center}
		\includegraphics[width=0.95\linewidth]{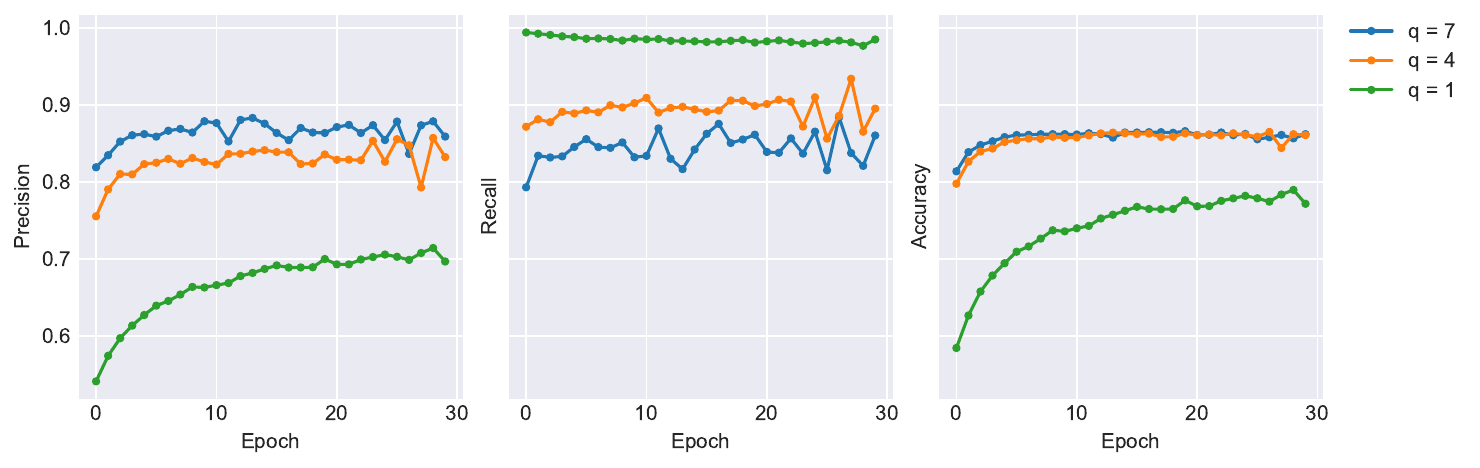}
	\end{center}
	\caption{The effect of hyperparameter $q$ on the Precision, Recall and Accuracy. }
	\label{fig:q_effect}
\end{figure*}

For the Multi-output classification with the \ac{CoTM}, the hyperparameter $q$
is extremely important, and can massively impact the performance of the model.
The hyperparameter $q$ decides the probability of a classes with false labels receiving the Type II feedback. A higher value of $q ( >1)$ means that, more classes with false labels
will receive the Type II feedback. Because of this, the clauses are able to
better distinguish between the patterns for different classes. Therefore, a
higher value of $q$ leads to higher precision and lower recall, while a lower
value of $q$ leads to higher recall and lower precision. Fig.~\ref{fig:q_effect} shows the effect of $q$ on different metrics for the CelebA
dataset. Interestingly, this hyperparameter does not have any significant effect for
multi-class classification.

\section{Conclusion}\label{sec:conclusion}

This paper address the challenge of interpreting convolutional clauses in
Tsetlin Machines for large-scale, image classification problems. We propose
novel methodologies for generating both local interpretations of individual
predictions and global class representations that aggregate important patterns
across classes. Our key contributions include: (1) a local interpretation
algorithm that maps activated clauses back to input features, providing direct
traceability superior to heatmap-based methods; (2) a global class
representation methodology using patch counting that reveals class-specific
patterns and dataset biases.

Experimental results show that our Convolutional CoTM achieves competitive
performance (98.5\% accuracy on MNIST, 86.56\% F1-score on CelebA compared to
88.07\% for ResNet50) while maintaining interpretability. Unlike CAM-based
methods that provide only heatmaps of important regions, our approach enables
direct mapping of predictions to input features, offering superior transparency
for critical applications.

The proposed methodologies successfully bridge the gap between interpretability
and performance in large-scale image classification, demonstrating that
transparent machine learning models can achieve competitive results on complex,
multi-channel datasets. The global class representations effectively highlight
inter-class differences and reveal dataset biases, providing valuable insights
for model validation and bias detection.

\subsection*{Limitations and Future Work}

Our experiments reveal that \ac{TM} models are highly sensitive to class
imbalance, particularly in multi-label datasets, limiting their applicability
to the full CelebA dataset. Future work will explore data balancing techniques
and robust training methodologies to handle imbalanced datasets. Additionally,
we will investigate approaches that leverage the learning dynamics of the
\ac{TM} to fix the imbalance issue. We also plan to extend our interpretation
methods to other TM variants including \ac{TM} Composits and investigate their
applicability to other domains such as natural language processing.

% Future work - How to make the interpretation independent of location. - local bound clauses kills generalization,
% making it impossible to learn objects, in the current scenario.

% \section*{Acknowledgment}
%
% The preferred spelling of the word ``acknowledgment'' in America is without an
% ``e'' after the ``g''. Avoid the stilted expression ``one of us (R. B. G.)
% thanks $\ldots$''. Instead, try ``R. B. G. thanks$\ldots$''. Put sponsor
% acknowledgments in the unnumbered footnote on the first page.

\bibliographystyle{IEEEtran}
\bibliography{references}

% \vspace{12pt}
%
% \color{red}
%
% IEEE conference templates contain guidance text for composing and formatting
% conference papers. Please ensure that all template text is removed from your
% conference paper prior to submission to the conference. Failure to remove the
% template text from your paper may result in your paper not being published.

\end{document}